\documentclass[letterpaper,10pt,conference]{ieeeconf}  

\usepackage{amsmath}
\usepackage{bbm}
\usepackage[pagebackref,breaklinks,colorlinks,citecolor=cvprblue]{hyperref}
\usepackage[dvipsnames]{xcolor}

\newcommand{\real}{\mathbb{R}}
\newcommand{\figref}[1]{Fig.~\ref{#1}}
\newcommand{\tabref}[1]{Table~\ref{#1}}

\usepackage[dvipsnames]{xcolor}

\usepackage{url}

\usepackage{algorithm}
\usepackage{algorithmic}

\usepackage[utf8]{inputenc} 
\usepackage[T1]{fontenc}    
\usepackage{multirow}
\usepackage{tikz}

\usepackage{booktabs}
\usepackage[font=small]{caption}

\usepackage{url}            
\usepackage{booktabs}       
\usepackage{amsfonts}       
\usepackage{nicefrac}       
\usepackage{microtype}      
\usepackage{xcolor}         
\usepackage{amsmath}

\IEEEoverridecommandlockouts                              

\title{\LARGE \bf
All-day Depth Completion
}

\author{Vadim Ezhov$^{1*}$, Hyoungseob Park$^{1*}$, Zhaoyang Zhang$^{1*}$, Rishi Upadhyay$^{2}$, Howard Zhang$^{2}$, \\ Chethan Chinder Chandrappa$^{2}$, Achuta Kadambi$^{2}$, Yunhao Ba$^{2}$, Julie Dorsey$^{1}$, Alex Wong$^{1}$
\thanks{*Equal contribution}
\thanks{$^{1}$Vadim Ezhov, Hyoungseob Park, Zhaoyang Zhang, Julie Dorsey and Alex Wong are with the Department of Computer Science, Yale University, 
        CT 06520, USA,
        {\tt\small \{vadim.ezhov,hyoungseob.park,zhaoyang.zhang, julie.dorsey,alex.wong\}@yale.edu}}%
\thanks{$^{2}$Rishi Upadhyay, Howard Zhang, Chethan Chinder Chandrappa, Yunhao Ba and Achuta Kadambi are with UCLA, 
        CA 90095, USA
        {\tt\small \{rishiu,chinderc, yhba\}@ucla.edu,hwdz15508@g.ucla.edu, achuta@ee.ucla.edu}}%
}

\begin{document}

\maketitle
\thispagestyle{empty}
\pagestyle{empty}

\begin{abstract}
We propose a method for depth estimation under different illumination conditions, i.e., day and night time. As photometry is uninformative in regions under low-illumination, we tackle the problem through a multi-sensor fusion approach, where we take as input an additional synchronized sparse point cloud (i.e., from a LiDAR) projected onto the image plane as a sparse depth map, along with a camera image. The crux of our method lies in the use of the abundantly available synthetic data to first approximate the 3D scene structure by learning a mapping from sparse to (coarse) dense depth maps along with their predictive uncertainty -- we term this, SpaDe. In poorly illuminated regions where photometric intensities do not afford the inference of local shape, the coarse approximation of scene depth serves as a prior; the uncertainty map is then used with the image to guide refinement through an uncertainty-driven residual learning (URL) scheme. The resulting depth completion network leverages complementary strengths from both modalities -- depth is sparse but insensitive to illumination and in metric scale, and image is dense but sensitive with scale ambiguity. SpaDe can be used in a plug-and-play fashion, which allows for 25\% improvement when augmented onto existing methods to preprocess sparse depth. We demonstrate URL on the nuScenes dataset where we improve over all baselines by an average 11.65\% in all-day scenarios, 11.23\% when tested specifically for daytime, and 13.12\% for nighttime scenes.

\end{abstract}

\section{Introduction}
Three-dimensional (3D) reconstruction, i.e., depth estimation, facilitates spatial tasks such as virtual and augmented reality, and autonomous navigation and manipulation. Existing works, from monocular to multi-view depth estimation, are largely trained and tested on well-illuminated environments. But when transferred to low-illumination scenarios, i.e. nighttime, the performance of these methods drops drastically due to a domain gap -- a covariate shift in the photometric intensities induced by the change in lighting conditions -- and in the absence of light, depth cannot be estimated from solely image-based methods. Efforts to reduce the performance gap mainly focus on re-balancing the training dataset by introducing additional images captured in the low-illumination environments. As manual curation of datasets with ground truth depth is expensive, existing training sets are augmented with images synthesized through means including, but not limited to, synthetic rendering or image-to-image translation through a generative model. However, rendering may introduce a synthetic to real domain gap, and image-to-image translation may introduce artifacts.

Counter to current trends, we instead investigate the use of a sparse range sensor, i.e., LiDAR, in addition to a camera, with the aim to robustly reconstruct the 3D scene structure under different lighting conditions, i.e., well-lit daytime and lowly-illuminated nighttime, for all-day depth estimation. Specifically, our approach estimates ego-centric dense depth maps from synchronized images and sparse point clouds projected onto the image plane, e.g., sparse depth maps. Nonetheless, the process of image-guided sparse point cloud (depth) completion is still ill-posed for each pixel without a measured point and susceptible to the photometric covariate shift. But while the point cloud is sparse, we have strong priors about the natural shapes of objects populating the 3D scene based on the configuration of the sparse points. This prior can serve as a form of inductive bias for depth estimation in regions where photometry is uninformative, i.e., poorly lit. Hence, we propose to approximate the 3D scene structure, from the sparse points, as a dense depth map and additionally estimate its predictive uncertainty to gauge the reliability of the approximated dense depth map. To this end, we leverage the abundance of publicly available synthetic data, where high quality ground truth can be used as supervision.

Once learned, our sparse to dense (SpaDe) approximation module can be used in a plug-and-play fashion by preprocessing sparse depth maps for existing methods, pretrained on daytime scenes, to extend them to all-day scenarios. Using plug-and-play with improved versions of SpaDe also improves overall performance. In another mode, existing models can be augmented with SpaDe, where its outputs (depth and uncertainty) can be adaptively fused with those of the downstream model via an uncertainty-driven residual learning (URL) scheme. We find that URL is not only performant, but also only adds a negligible amount of run-time (8 ms) and memory usage ($\approx$1GB) while enabling all-day depth estimation. We evaluate our approach on three recent depth completion methods on the nuScenes \cite{nuscenes} dataset and improve by an average of 11.23\% in day, 13.12\% in night and 11.65\% overall.

Our contributions are 
(i) a light-weight plug-and-play network (SpaDe) to approximate dense depth with predictive uncertainty from sparse points, and (ii) an uncertainty-driven residual learning scheme that alleviates existing models from the need to learn depth from scratch by leveraging SpaDe as an inductive bias. (iii) Plug-and-play with SpaDe is forward-compatible; future (better) versions of SpaDe can further improve results in a seamless integration manner. To the best of our knowledge, this is the first approach to address all-day depth estimation from image and sparse range fusion.

\vspace{-0.5em}
\section{Related Works}

\textbf{Supervised depth completion} learns a mapping from images and corresponding sparse depth maps to dense depth maps with ground truth. Earlier works undertook approaches of compressing sensing \cite{chodosh2019deep} and approximating morphological operators \cite{matrin2018morphological}. A line of works catered to sparse data by altering network operations \cite{uhrig2017sparsity,eldesokey2018propagating,huang2020hmsnet}, and extending architectures \cite{chen2019learning}, \cite{yang2019dense}. Employing RGB guidance, \cite{ma2018selfsupervised} proposed early fusion after initial convolution. \cite{huang2020hmsnet} used encoder features of concatenated modalities to upsample the sparse depth map. \cite{hu2020PENet} extended this approach by two-stage sequential fusion. \cite{li2020multi} used multi-scale cascade hourglass network. \cite{park2020non} implemented non-local spatial propagation, improving over fixed-local methods \cite{cheng2020cspn++}. Several works incorporated auxiliary data in form of confidence maps \cite{van2019sparse} and uncertainty estimations \cite{eldesokey2018propagating,Eldesokey_2020_CVPR,qu2020depth,qu2021bayesian}.

\textbf{Unsupervised depth completion} \cite{ma2019self, wong2021learning, wong2021adaptive,wong2020unsupervised,shivakumar2019dfusenet, yang2019dense} learns depth by minimizing: (1) sparse depth reconstruction and (2) photometric error between the original image and its reconstruction from other views of the same scene. \cite{ma2019self} estimates transformation between consecutive frames using Perspective-n-Point \cite{lepetit2009epnp} and RANSAC \cite{FISCHLER1987726}. \cite{yang2019dense} used separate network to capture depth prior given an image. \cite{wong2021learning} also learned to approximate dense depth from sparse depth maps, but does not consider uncertainty, nor low illumination scenarios.  \cite{wong2021unsupervised} proposed a calibrated backprojection layer. \cite{jeon2022struct} used line feature from visual SLAM. \cite{yan2023desnet} decouples structure and scale. 

All of the above are designed for well-illuminated scenarios. Specifically, unsupervised methods rely on the photometric reconstruction loss, which requires temporal consistency with minimal occlusions in consecutive frames without specular reflections; current unsupervised methods cannot be trained for nighttime scenes. Thus, we explore supervised learning paradigm for all-day depth estimation.

\textbf{All-day and nighttime depth estimation} remain  challenging due to a loss of photometric information (low signal to noise) from low illumination and inconsistent exposure. \cite{Wang2021RegularizingNW} used image enhancement and adaptive masking nighttime scenes. \cite{zheng2023steps} extended the approach to all-day estimation by jointly learning enhancement module. Other works bridged domain gap using image translation \cite{sharma2020nighttime} and discriminative learning \cite{nightdepthADFA} models. \cite{liu2021self} instead proposed extracting view-invariant and variant features with an encoder for each domain. \cite{vankadari2023sun} demonstrated illumination-invariant photometric loss, compensating for various exposure and motion by image denoising and predicting per-pixel residual flow map. \cite{Kim2018MultispectralTN, Lu2021AnAO} also relied on alternative modality less affected by illumination -- thermal images. \cite{Lu2021AnAO} estimated depth directly from one thermal image while training with RGB images.

Unlike single-modality (monocular) depth estimation, we fuse RGB camera images and synchronized sparse depth maps from LiDAR, which is invariant to illumination changes. We leverage the complementary strengths of these modalities to perform all-day depth estimation without the need for enhancement or image-to-image translation during training.

\vspace{-0.5em}
\section{Method}

\textbf{Motivation.} 
Daytime and nighttime images exhibit significant difference in illumination, posing a challenge for depth estimation. 
To address this 
we investigate the efficacy of multi-sensor fusion:  
\textit{Image sensors} (i.e. CMOS sensors in camera) capture dense 2D projections of the 3D scene -- photometry is naturally sensitive to illumination.
While daytime images typically present distinct object appearances, which allows one to infer object shapes, nighttime images are often presented with low illumination and photometric disturbances (e.g. low signal to noise).
Conversely, \textit{range sensors} (e.g. LiDAR, radar) capture a sparse point cloud of the 3D scene. 
The inherent robustness of range sensors 
under different lighting conditions motivates us to exploit it as an additional modality for depth estimation. While range sensors can resolve depth, their measurements are often sparse -- leading to trade-offs between the image (dense, but sensitive to illumination) and sparse range modalities.
To this end, we present a method to cohesively integrate different sensor modalities to estimate depth under ``all-day'' (day and nighttime) scenarios.

\textbf{Formulation.} Given datasets $\mathcal{D}_\text{day}$ and $\mathcal{D}_\text{night}$ comprised of daytime and nighttime scenes, respectively, we aim to estimate depth under all-day scenarios i.e. $\mathcal{D}_\text{all-day} = (\mathcal{D}_\text{day} \cup \mathcal{D}_\text{night}$). We assume data samples $(I, z, d^*) \in \mathcal{D}_\text{all-day}$, where $I \in \real^{3 \times H \times W}$ denotes an RGB image, $z \in \real_+^{H \times W}$ a synchronized point cloud projected onto the image plane as a sparse depth map, and $d^* \in \real_+^{H \times W}$ the ground truth depth. Note: in all-day depth estimation, aside from the variation in illumination between daytime and nighttime images, which degrades performance, there also exists an imbalance in the data for each lighting condition to further exacerbate the errors.

\begin{figure}
    \centering
    \includegraphics[width=0.70\linewidth]{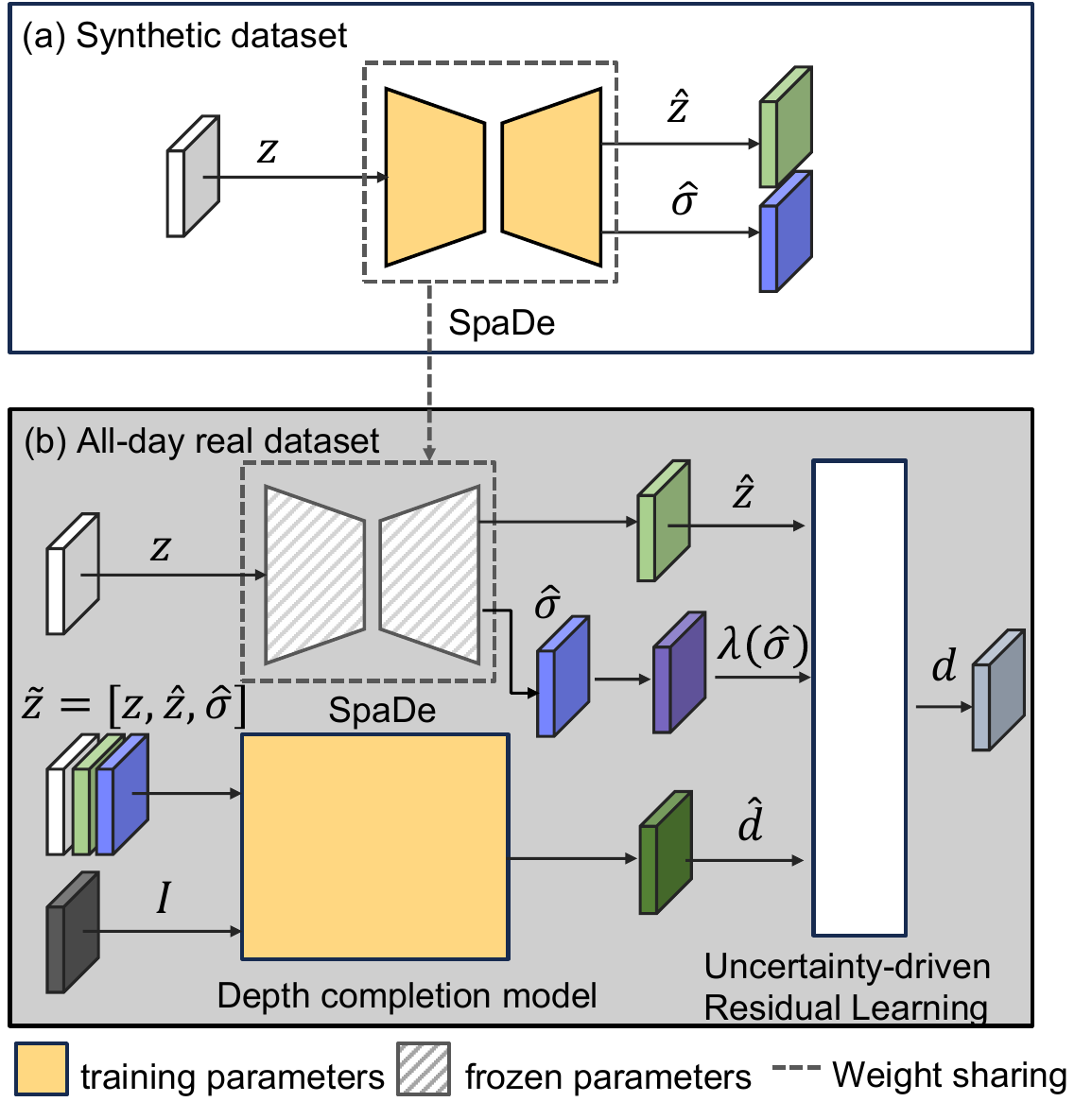}
    \vspace{-0.0em}
    \caption{
    \textbf{An Illustration of Uncertainty-driven Residual Learning}. (a) Our Sparse-to-Dense module (SpaDe) is trained on synthetic data to approximate dense depth from sparse points. (b) SpaDe is used as an inductive bias in Uncertainty-driven Residual Learning (URL), and the depth estimation model is trained to adaptively refine the approximated depth based on the estimated log uncertainty. SpaDe can also be used in a plug-and-play manner to enable all-day depth estimation for a pretrained depth estimator without training.
    }
    \vspace{-2em}
    \label{fig:overview}
\end{figure}

\begin{figure*}[h!]
    \centering
    \includegraphics[width=0.85\linewidth]{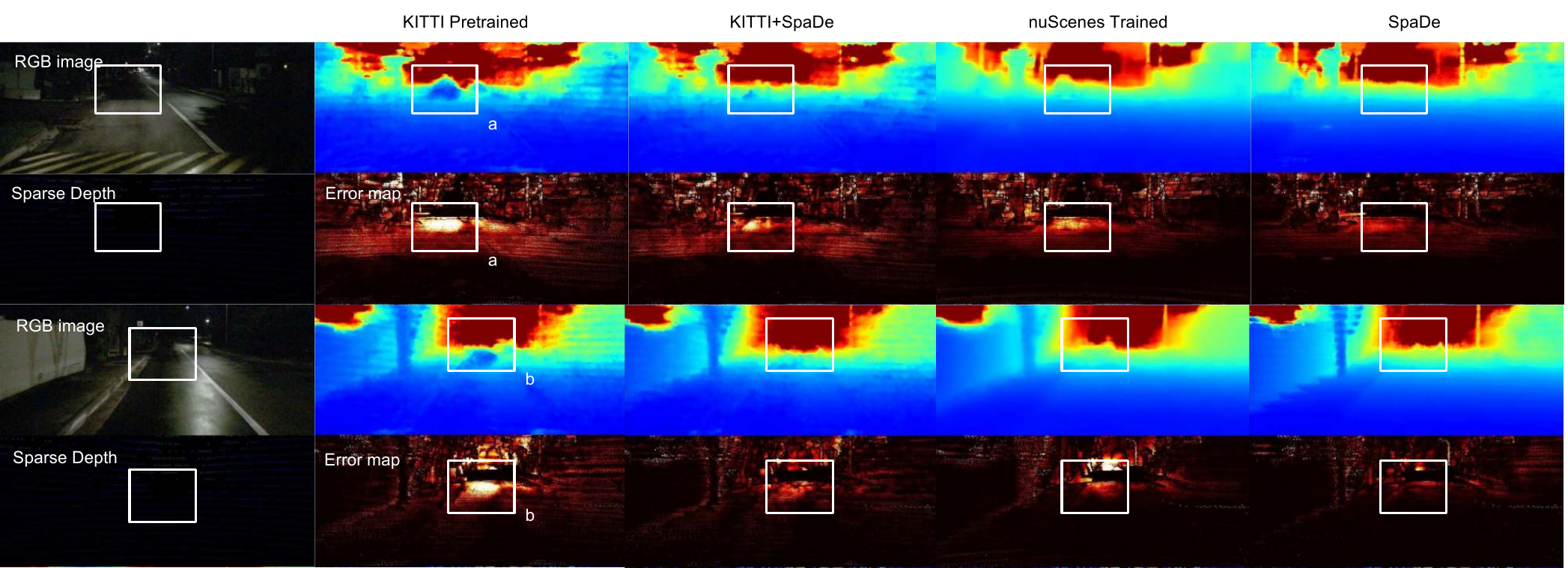}
    \vspace{-1mm}
    \caption{
    \textbf{Ablation Study on nuScenes} Predictions from SpaDe serve as strong inductive bias for downstream depth completion models. Augmenting KITTI pretrained models with SpaDe improves estimates in regions where photometry is uninformative (highlighted). }
    \label{fig:ablation}
    \vspace{-4mm}
\end{figure*}

We propose first to learn a deep neural network $f_\text{SpaDe}$ from synthetic data to approximate the coarse 3D scene structure as a dense depth map and its predictive uncertainty: $[\hat{z}, \hat{\sigma}] = f_\text{SpaDe}(z)$ where $\hat{z} \in \real_+^{H \times W}$ refers to the predicted depth map and $\hat{\sigma} \in \real^{H \times W}$ to the uncertainty.
Once trained, $f_\text{SpaDe}$ can be frozen (see \figref{fig:overview}-(a)) and augmented onto existing pretrained depth completion networks $f_\text{DC}$ to enable all-day depth estimation. We hypothesize that the dense depth predicted by $f_\text{SpaDe}$ will serve as a strong prior to bias predictions in image regions that are uninformative (\figref{fig:ablation}).

In its ``plug-and-play'' mode, $f_\text{SpaDe}$ is used as a preprocessing step to densify the input sparse depth map to a pretrained depth completion network $f_\text{DC}$. Sparse depth $z$ and predicted depth $\hat z$ are combined into a single input depth map $\tilde{z} \in \real_+^{H \times W}$ where high uncertainty regions in $\hat{z}$ are first filtered out based on $\hat{\sigma}$; any location in $\hat{z}$ where there exists $z$ is substituted with its value: $\tilde{z} = \mathbbm{1}_z \cdot z  + (1 - \mathbbm{1}_z) \cdot \mathbbm{1}_{\hat{\sigma}} \cdot \hat{z}$, where $\mathbbm{1}_z, \mathbbm{1}_{\hat{\sigma}} \in \{0, 1 \}^{H \times W}$ are indicators for positions where positive $z$ exists and $\hat{\sigma}$ is below an empirically chosen threshold $\tau = 5.0$, respectively. The output depth $d \in \real_+^{H \times W}$ is thus $d = f_\text{DC}(I, \tilde{z})$ when using SpaDe in plug-and-play.

In a more performant mode, one can train $f_\text{DC}$ while freezing $f_\text{SpaDe}$. Here, $f_\text{SpaDe}$ serves again as a preprocessing step, except that we do not mix $z$ and $\hat{z}$, but concatenate them together with uncertainty. For ease of notation, we overload $\tilde{z} = [z, \hat{z}, \hat{\sigma}]$. To alleviate the burden of learning depth from scratch, one may leverage $f_\text{DC}$ to correct more uncertain regions, allowing one to dedicate model capacity to learning the residual $\hat{d}$ by minimizing typical supervised loss w.r.t. ground truth $d^*$ on $\mathcal{D}_\text{all-day}$. The final output is attained with uncertainty-weighted scheme $d = \lambda(\hat{\sigma}) \hat{z} + (1-\lambda(\hat{\sigma}))\hat{d}$, where balancing factor $\lambda(\hat{\sigma})$ grants greater contribution to $\hat{d}$ in higher uncertainty regions. This lends to an uncertainty-driven residual learning scheme (see \figref{fig:overview}-(b)) where $f_\text{DC}$ refines $\hat{z}$ with $\hat{d}$ based on predictive uncertainty $\hat{\sigma}$. 

\vspace{-0.em}
\subsection{Learning to approximate dense depth from sparse range}
Given a synthetic dataset with training samples $(z, d^*_{\text{syn}}) \in \mathcal{D}_\text{syn}$, where $d^*_\text{syn}$ denotes the rendered ground truth, we propose a light-weight convolutional encoder-decoder, Sparse-to-Dense network (SpaDe), $[\hat{z}, \hat{\sigma}] = f_\text{SpaDe}(z)$, which not only approximates the dense depth $\hat{z}$, but also estimates its log uncertainty $\hat{\sigma}$, from sparse depth map $z$. To learn $f_\text{SpaDe}$, we leverage the high quality, dense ground truth that can be readily obtained in synthetic datasets, sourced from depth buffers in 3D rendering engines, and minimize an L2 loss:
\begin{equation}
    \mathcal{L}_{\text{SpaDe-z}}(\hat{z}, d_\text{syn}^*) = ||\hat{z} - d_\text{syn}^*||^2_2.
\end{equation}
While the range sensor is robust to illumination changes, the approximated depth may contain erroneous regions, given that it is solely predicted from sparse points. To quantify the reliability of each predicted point, we additionally predict the log-Gaussian uncertainty $\hat{\sigma}$ as a dense map $\hat{\sigma} \in \real^{H \times W}$. 

The log uncertainty loss function aims to learn the predictive uncertainty $\hat{\sigma} \in {\mathbb{R}^{H\times W}}$ of the approximated depth, assuming  the uncertainty of $\hat{z}$ follows a Gaussian distribution,
\begin{equation}
    \mathcal{L}_{\text{SpaDe}-{\sigma}}(\hat{z}, d_\text{syn}^*) = \frac{1}{2}\left( \frac{\hat{z}-d_\text{syn}^*}{e^{\hat{\sigma}}} \right)^2 + \hat{\sigma}.
\end{equation}

\begin{figure}
    \centering
    \includegraphics[width=1.0\linewidth]{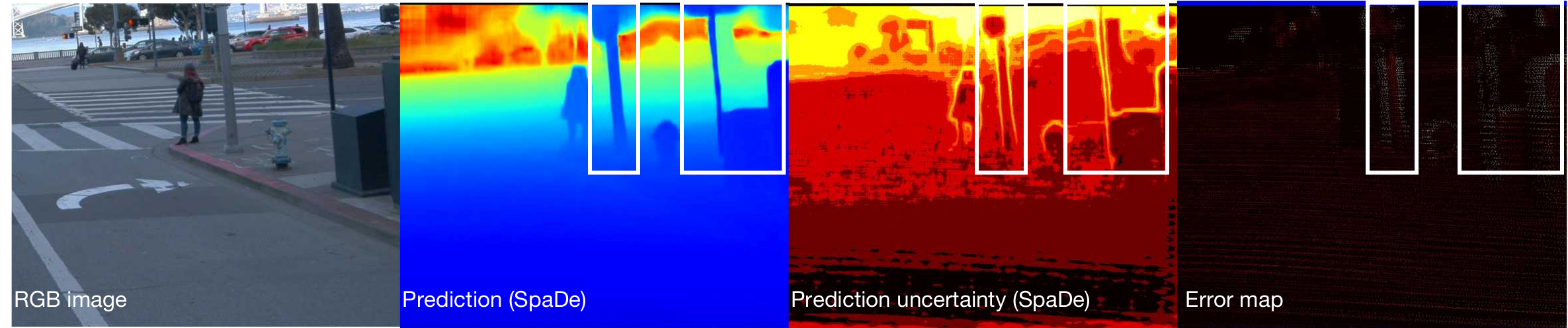}
    \vspace{-4mm}
    \caption{
    \textbf{Sparse-to-Dense module (SpaDe) on real dataset}. The boxes highlight the alignment between SpaDe's predictive uncertainty and depth discontinuity regions that often erroneous. The uncertainty aligns well with error when compared to ground truth.}
    \label{fig:uncertainty}
    \vspace{-6mm}
\end{figure}

In practice, we train for depth first, and then train the uncertainty decoder while freezing encoder and depth decoder.
To show the efficacy of SpaDe as a inductive bias, we visualize the prediction, uncertainty and error on Waymo (\figref{fig:uncertainty}).

\subsection{Uncertainty-driven residual learning (URL)}

Given available datasets, one may also augment a depth completion network $f_{\text{DC}}$ with SpaDe $f_\text{SpaDe}$ and train it to learn the residual $\hat{d}$ of the predicted depth map $\hat{z}$, which re-purposes the downstream $f_{\text{DC}}$ as a refinement module for $\hat{z}$. To leverage $f_\text{SpaDe}$ as an inductive bias, the refinement can be conducted adaptively, where the weighting function $\lambda(\hat{\sigma})$ assigns residuals $\hat{d}$ with larger weight for higher $\hat{\sigma}$ (more freedom to modify $\hat{z}$) and likewise lower weight for lower $\hat{\sigma}$:

\begin{equation}
    \lambda(\hat{\sigma}) = \frac{1}{1+e^{\alpha({\hat{\sigma}-\beta})}},
\end{equation}
where the $\alpha = 0.8$ and $\beta = 0$ are hyperparameters to calibrate the reliability of the respective models. Our uncertainty-driven residual learning (URL) scheme manifests as a linear combination balanced by $\lambda(\hat{\sigma})$. The output depth reads
\begin{equation}
    {d} = \lambda(\hat{\sigma}) \hat{z} + (1-\lambda(\hat{\sigma}))\hat{d},
\end{equation}
where $\lambda(\cdot)$ is the weighting function based on log uncertainty value. Intuitively, $\lambda(\cdot)$ guides the downstream $f_{\text{DC}}$ to prioritize the reduction of errors in regions exhibiting high uncertainty.

To learn $f_{\text{DC}}$, we minimize a supervised loss on $\mathcal{D}_\text{all-day}$. While any depth completion base model can be seamlessly integrated into our framework, we consider three models for $f_\text{DC}$ in URL, where each model minimizes the loss function specified in their respective paper, which follows the form:
\begin{equation}
    \mathcal{L}_{\text{sup}} = || d - d^* ||_p,
\end{equation}
where $p$ denotes the L-p norm for the loss, either L1 or L2. 

\begin{figure*}[!t]
    \centering
    \includegraphics[width=0.65\linewidth]{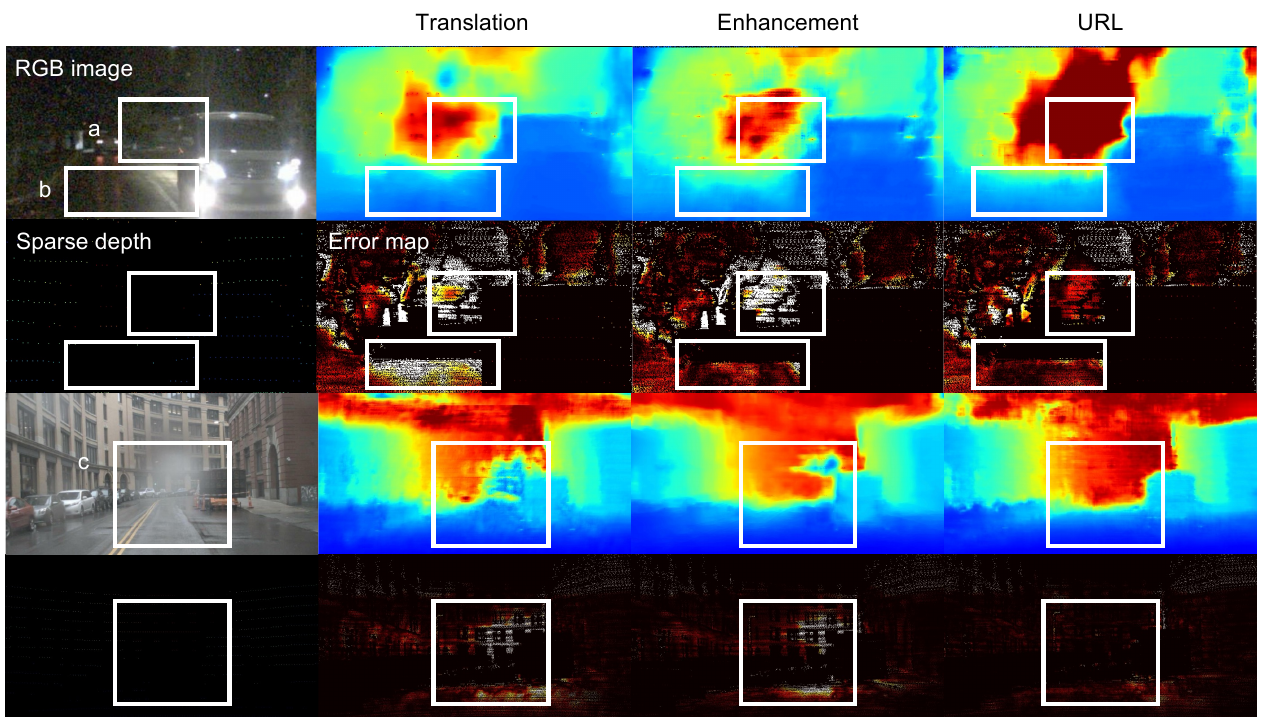}
    \caption{\textbf{Representative results of all-day depth estimation on nuScenes day and night images.} The region for detailed comparisons are highlighted by boxes. URL performs better on (a) low-illumination conditions, (b) missing sparse points and (c) depth discontinuity regions. }
    \label{fig:method_overview}
    \vspace{-4mm}
\end{figure*}

Additionally, as ground truth in real datasets are at most semi-dense, we also include a local smoothness regularizer on $d$ as part of the training objective. 
Specifically, local smoothness loss is computed on gradients of depth prediction in both horizontal and vertical directions, $\partial_X$ and $\partial_Y$, respectively. 
We denote the smoothness loss $\mathcal{L}_\text{sm}$ as follows:
\begin{equation}
    \mathcal{L}_{\text{sm}} = \frac{1}{|\Omega|} \sum_{ x \in \Omega} I_X(x)|\partial_X {d}(x)| + I_Y(x)| \partial_Y {d}(x)|,
    \label{eqn:loss_smoothness}
\end{equation}
where $I_X(x) = e^{-|\partial_X I(x)|}$, $I_Y(x) = e^{-|\partial_Y I(x)|}$ and $\Omega$ denotes image domain.
The total loss function for $f_\text{DC}$ reads:
\begin{equation}
    \mathcal{L}_{\text{total}} = w_{\text{sm}}\mathcal{L}_{\text{sm}} + w_{\text{sup}}\mathcal{L}_{\text{sup}},
\end{equation}
where $w_{\text{sup}}$ and $ w_{\text{sm}}$ are the weights for each loss term.

\section{Experiments and results}
We consider three recent depth completion models: ENet \cite{hu2020PENet}, MSG-CHN\cite{li2020multi} and CostDCNet \cite{kam2022costdcnet}. Each is evaluated using KITTI \cite{geiger2012we} metrics (MAE, RMSE, iMAE, iRMSE)  under daytime, nighttime, and all-day scenarios.

\subsection{Datasets}
\noindent\textbf{nuScenes} \cite{nuscenes} is an outdoor dataset comprised of 1000 scenes from Boston and Singapore. Since each sensing modality captures the scene at a different frequency, the authors find keyframes, where the sensors timestamps are synchronized. The dataset contains around 40,000 annotated keyframes (around 40 samples per scene).
We use the original nuScenes train/val split (700 scenes for train, 150 for val).
Note: we use the 544x1600 bottom crop to validate the models.

\textbf{Waymo Open Dataset} \cite{sun2020scalability} consists of 1150 scenes from different illuminations (day/night/dawn). Each scene includes average 197 frames with high-quality synchronized LIDAR and RGB image. We used Waymo to train a second set of baselines to test the applicability of SpaDe and a ``future'' update to SpaDe (trained with more data) for plug-and-play. We show that SpaDe can even improve methods trained with images captured under different illumination conditions. 

\textbf{Virtual KITTI} (VKITTI) \cite{gaidon2016virtual} consists of 35 synthetic videos (5 cloned from the KITTI \cite{uhrig2017sparsity}, each with 7 variations in weather, lighting or camera angle) for a total of $1242\times375$ sized $\approx$17K frames. 
We only use the dense depth maps of VKITTI to train SpaDe. To acquire the sparse points, we imitate the sparse depth measurement of nuScenes.

\textbf{SYNTHIA} \cite{7780721} is a synthetic collection of urban scenes, rendered in a virtual city. Virtual world includes all fours seasons, with scenes captured under dynamic illumination conditions. We only utilize dense maps in conjunction with VKITTI to train SpaDe-V2.

\subsection{Implementation details} 

All models were trained using four NVIDIA RTX 3090 GPUs. MSGCHN, ENet and CostDCNet used 1, 2, and 4 GPUs respectively. SpaDe was trained on two NVIDIA RTX 3080 Ti GPUs. MSG-CHN took 32hrs with a batch size of 16. ENet and CostDCNet took 55 hours and 30 hours, respectively, with corresponding batch sizes of 12 and 24.

Note: the sparse depth map $z$, was passed to the geometric convolutional layer of ENet and to the 3-D encoder of CostDCNet as required by their methods. All models take in $\tilde{z} = [z, \hat{z}, \hat{\sigma}]$ as input to their respective 2-D depth encoder.

SpaDe was trained on VKITTI and later frozen for URL. We trained it for 30 epochs with a learning rate of 2e-4 to learn depth, and later froze the encoder and depth decoder for training uncertainty decoder for a total of 55 epochs with an initial learning rate of 2e-4, and reduce to 1e-4 and 5e-5 at epoch 25 and 40. For the baselines, MSGCHN was trained for 90 epochs with initial learning rate of 1e-3, reduced at 10th, 20th, and 25th epoch to 5e-4, 2e-4, 1e-4 respectively. ENet was trained for 50 epochs under similar schedule, while CostDcNet was trained for 85 epochs using 2e-4 learning rate. We use color jitter, random resize and crop, horizontal flip for augmentations and the crop size of 544x704.

\begin{table*}[t]
\scriptsize
\centering
\setlength\tabcolsep{3pt}
\caption{\textbf{Plug-and-play evaluation on nuScenes for daytime, nighttime, and all-day depth completion.} SpaDe improves models with plug-and-play and is also forward-compatible with SpaDe-V2. Best results are in \textbf{bold}, and second place results are \underline{underlined}.}
\vspace{0mm}
\resizebox{0.99\textwidth}{!}{
\begin{tabular}{cl cccc cccc cccc}
    \toprule
    &  & \multicolumn{4}{c}{nuScenes-Daytime} & \multicolumn{4}{c}{nuScenes-Nighttime} & \multicolumn{4}{c}{nuScenes-All}\\
    
    \midrule
    Method & {} & MAE & RMSE & iMAE & iRMSE & MAE & RMSE & iMAE & iRMSE & MAE & RMSE & iMAE & iRMSE \\
     
    \midrule
    SpaDe & VKITTI &1704.850 & 4280.270 & 4.321 & 8.765 & 1755.366 & 4335.351 & 4.457 & 10.391 & 1709.207 & 4282.599 & 4.453 & 8.951 \\
    SpaDe-V2 &VKITTI+Synthia & 1621.320 & 4060.945 & 4.230 & 8.582 & 1693.932 & 4198.802 & 4.507 & 10.244 & 1628.582 & 4074.733 & 4.258 & 8.748 \\
    \midrule
    \multirow{6}{*}{MSG-CHN~\cite{li2020multi}} 
    & KITTI Pretrained  &3301.149 & 6136.909& 10.848  & 16.275&3083.199 &5680.178 & 10.738&17.785 &3279.350 & 6091.228& 10.837&16.426 \\
    & +SpaDe & \underline{1697.258} &  \underline{4268.812} &   \underline{4.381}   & \underline{8.812} & \underline{1748.739} &  \underline{4323.876}  & \textbf{4.569}  &  \underline{10.343} & \underline{1700.709}	&\underline{4270.050}&	\underline{4.395}&	\underline{8.956} \\
    & +SpaDe-V2 & \textbf{1636.697} & \textbf{4070.550}  &  \textbf{4.415} & \textbf{8.753} &  \textbf{1706.174}  &  \textbf{4193.231} & \underline{4.634} & \textbf{10.292} & \textbf{1642.008}	&\textbf{4078.748}&	\textbf{4.432}&	\textbf{8.898}  \\
    \cmidrule{2-14}
    & Waymo Pretrained  & 1834.319 & 4241.108 & 7.279 & 14.610 & 1802.856 & 4195.771 & 7.425 & 16.218 &  1829.338&	4232.333&7.286&	14.756 \\
    & +SpaDe  & \underline{1675.269} &  \underline{4094.963} & \underline{6.123}  & \underline{12.199} & \underline{1767.207} & \underline{4238.647} & \underline{6.374} & \underline{13.722} &  \underline{1682.788}	&\underline{4105.236}&	\underline{6.142}& \underline{12.339} \\
    & +SpaDe-V2  & \textbf{1628.381} & \textbf{3976.477} & \textbf{6.082}  & \textbf{12.144} & \textbf{1739.883} & \textbf{4172.608} & \textbf{6.353} & \textbf{13.669} &  \textbf{1637.903}	& \textbf{3992.114} & \textbf{6.103} & \textbf{12.284} \\
    
    \midrule
    \multirow{6}{*}{ENet~\cite{hu2020PENet}} 
    & KITTI Pretrained  &7216.192 &11184.920 &34.710 &42.746 &8631.241 &12836.475 &37.107 & 46.974&7357.721 & 11350.104&  34.950& 43.169 \\
    & +SpaDe &\underline{1765.793} & \underline{4360.097} &  \underline{4.480} & \underline{8.832} & \underline{1884.805} & \underline{4477.324} & \underline{4.740}  & \underline{10.437} & \underline{1775.928}&	\underline{4367.460} &	\underline{4.502} &	\underline{8.984} \\
    & +SpaDe-V2 & \textbf{1748.582} & \textbf{4214.509} & \textbf{4.381} & \textbf{8.663} & \textbf{1875.735} & \textbf{4374.235} & \textbf{4.673} & \textbf{10.300} & \textbf{1759.549}	& \textbf{4226.267} & \textbf{4.406}	& \textbf{8.818}  \\
    \cmidrule{2-14}
   & Waymo Pretrained & 1909.895 & 4337.763 & 8.585 & 16.568 & 2073.584 & 4549.566 & 8.822 & 17.827 & 1924.354& 4354.606 & 8.600 & 16.677 \\
    & +SpaDe & \underline{1821.182} & \underline{4216.877} & \underline{7.683}  & \underline{14.775} &  \underline{1981.966} & \underline{4423.804} & \underline{7.925}  & \underline{16.004} & \underline{1835.439} & \underline{4233.353}&  \underline{7.700} & \underline{14.883}\\
    & +SpaDe-V2& \textbf{1776.277} & \textbf{4102.502} & \textbf{7.649} & \textbf{14.715} & \textbf{1956.100} & \textbf{4364.737}  & \textbf{7.900} & \textbf{15.955} & \textbf{1792.483}	& \textbf{4124.623} & \textbf{7.666}	& \textbf{14.824}\\
    
    \midrule
    \multirow{6}{*}{CostDCNet~\cite{kam2022costdcnet}} 
    & KITTI Pretrained & 2296.107 &4962.606 &	7.342&13.074& 2409.224&	5249.870 &	7.235 &	13.533 & 2305.123&4986.370 &7.324	&13.107   \\
        & +SpaDe & \underline{1796.246}	&\underline{4318.937}&	\textbf{5.174}&	\underline{9.735}&	\underline{1871.708}&	\underline{4463.116}&	\underline{5.082}&	\underline{10.975}&	\underline{1801.996}&	\underline{4329.036}&	\textbf{5.16}&	\underline{9.849} \\
    & +SpaDe-V2 & \textbf{1745.836} &\textbf{4136.050}&	\underline{5.227}	&\textbf{9.697}&\textbf{1706.399} &	\textbf{4194.396} &	\textbf{4.634} &\textbf{10.295}	&\textbf{1740.146}&	\textbf{4137.749}&	\underline{5.162}&	\textbf{9.747} \\
    \cmidrule{2-14}
    & Waymo Pretrained & 1955.058	&4415.014&	6.957&	13.001&	2061.947&	4681.097&	7.099&	14.500&	1963.792&	4437.207&	6.964&	13.138 \\
    & +SpaDe & \underline{1797.240}& \underline{4210.430}	&\underline{6.345}&	\underline{12.127}&	\underline{1872.100}&	\underline{4375.482}&	\underline{6.494}&	\underline{13.603}&	\underline{1802.929}&	\underline{4222.725}& \underline{6.354}	& \underline{12.262} \\
    & +SpaDe-V2 & \textbf{1744.971}& \textbf{4095.855} &\textbf{6.300} & \textbf{12.059} &\textbf{1824.424}& \textbf{4290.672}& \textbf{6.458} &\textbf{13.552}&	\textbf{1751.171}&	\textbf{4111.241}&	\textbf{6.310} &\textbf{12.196} \\
    
    \bottomrule
    
\end{tabular}
    } 
\vspace{-3mm}
\label{tab:experiments:plug_and_play}
\end{table*}

\begin{table*}[t]
\scriptsize
\centering
\setlength\tabcolsep{2pt}
\caption{\textbf{Evaluation on nuScenes for daytime, nighttime, and all-day depth completion.} Best results are in \textbf{bold}, and second place results are \underline{underlined}. Despite not having extra training data or image enhancement, SpaDe and URL improve all three baselines by 11.65\%.}
\vspace{-1mm}
\resizebox{0.99\textwidth}{!}{
\begin{tabular}{cl c cccc cccc cccc}
    \toprule
    &   & \multicolumn{4}{c}{nuScenes-Daytime} & \multicolumn{4}{c}{nuScenes-Nighttime} & \multicolumn{4}{c}{nuScenes-All}\\
    
    \midrule
    Architecture  & Method & MAE & RMSE & iMAE & iRMSE & MAE & RMSE & iMAE & iRMSE & MAE & RMSE & iMAE & iRMSE \\
    \midrule 
  
    \multirow{5}{*}{MSG-CHN~\cite{li2020multi}}  
    & Baseline & 1674.465 & 3926.396 & 5.749 & 10.842 & 2068.051 & 4470.071  & 6.801 & 13.079 & 1713.830 &3980.773  & 5.854 & 11.066 \\
    &   +Translation &  1612.436 & 3837.921 & 5.650 & 13.066 & 1993.668 & 4564.014 & 6.066 & 12.290 &1650.566 &3910.543 &5.692 &12.988 \\
    &  +Enhancement & 1723.677  & 4345.944 & 4.519 & 8.747 & 2223.441 & 5118.473 & 6.049 & 12.37 & 1773.619 &  4423.110& 4.672 & 9.110 \\
    &  +SpaDe  & \underline{1370.595} & \underline{3550.412} & \underline{4.086} & \underline{8.102} & \underline{1650.553} & \underline{3875.170} & \underline{4.838} & \textbf{10.211} & \underline{1387.848} & \underline{3563.635} & \underline{4.073} & \underline{8.187} \\
    &  +URL  & \textbf{1332.280} & \textbf{3478.409} & \textbf{3.863} & \textbf{7.825} & \textbf{1521.641} & \textbf{3693.705} & \textbf{4.716} & \underline{10.346} & \textbf{1343.235} & \textbf{3485.414} & \textbf{3.909 }& \textbf{8.046} \\
    \midrule
    \multirow{5}{*}{ENet~\cite{hu2020PENet}}
    &  Baseline & 1338.449 & 3389.611 & 3.968 & 7.713 &1611.057 &3706.047 &5.322 &10.945 &1358.298 & 3406.133 &  4.081 &8.046 \\
    & +Translation & 1378.031 & 3449.752 & 4.582 & 8.740 &1581.408 & 3644.441  &  5.773& 11.716 & 1397.238& 3473.109&4.711 &9.052 \\
    & +Enhancement & 1314.127 & 3351.362 & 3.997 & 7.831 & 1771.770 & 3913.751 & 6.616 & 13.592 & 1359.865 & 3407.534 & 4.259 & 8.407\\
    & +SpaDe  &\underline{1280.334} & \textbf{3297.927} & \underline{3.932} & \underline{7.656} & \underline{1578.806} & \underline{3634.643} &  \underline{5.176} &\underline{10.579} & \underline{1310.143} & \textbf{3331.935} & \underline{4.057} & \underline{7.951}\\
    & +URL  & \textbf{1216.147} & \underline{3339.076} & \textbf{3.559} & \textbf{7.337} & \textbf{1439.265} & \textbf{3576.720} & \textbf{4.506} & \textbf{9.969} & \textbf{1238.682} & \underline{3363.078} & \textbf{3.655} & \textbf{7.602} \\
    \midrule
    \multirow{5}{*}{CostDCNet~\cite{kam2022costdcnet}}
    & Baseline &1221.396&3326.633&3.896&7.879&1451.260&3608.133&4.857&10.793&1243.161&3351.456&3.988&8.163  \\
    &  +Translation & 1202.946&3334.581&3.666&7.608&\underline{1404.093}&3623.874&\underline{4.239}&\underline{9.671}&1221.858&3360.176&3.72&7.807 \\
    & +Enhancement &1202.613&3344.348&3.765&7.882&1413.974&3571.865&4.934&11.138&1222.546&3363.755&3.878&8.2 \\
    & +SpaDe  & \underline{1200.852}&\underline{3318.437}&\underline{3.592}&\underline{7.471}&\textbf{1363.872}&\textbf{3554.628}&\textbf{4.145}&\textbf{9.643}&\underline{1215.953}&\underline{3338.738}&\underline{3.644}&\underline{7.681}  \\
    & +URL   & \textbf{1196.157}&\textbf{3280.429}&\textbf{3.542}&\textbf{7.402}&1405.163&\underline{3559.219}&4.474&10.238&\textbf{1215.861}&\textbf{3305.028}&\textbf{3.632}&\textbf{7.678} \\
    \bottomrule
    
\end{tabular}
} 
\vspace{-5mm}
\label{tab:experiments:outdoor}
\end{table*}

\noindent\textbf{Baselines.} We consider  KITTI \cite{uhrig2017sparsity} pretrained models of ENet \cite{hu2020PENet}, MSG-CHN \cite{li2020multi} and CostDCNet \cite{kam2022costdcnet} as baselines i.e. zero-shot generalization. Note that KITTI does not contain nighttime scenes. We also train baselines on Waymo to show that SpaDe can yield improvements even if a model was pretrained with nighttime imagery. Additionally, we train each method on the original nuScenes dataset, which contains an imbalanced day and night time scenes.
We further consider two adaptations of the previously mentioned baselines with image translation and image enhancement respectively.

\textbf{Image-to-image translation.} Following \cite{nightdepthADFA,sharma2020nighttime}, we train depth completion models on the data with extra images translated by day-to-night translation network (\tabref{tab:experiments:outdoor}, marked with ``+Translation''). Due to limited number of night-time images in nuScenes \cite{nuscenes}, the translation network is first pretrained on BDD \cite{yu2020bdd100k}, then finetuned on nuScenes.

\textbf{Enhancement baseline.} Following \cite{zheng2023steps}, we train our baselines with the SCI image enhancement module \cite{Ma_2022_CVPR}. SCI estimates the illumination stage by stage and generate a corresponding calibrated residual map, with which the image is enhanced. Although in \cite{zheng2023steps} they jointly train the depth estimation network and the image enhancement module, we directly use the pretrained image enhancement module and freeze it during our training of baselines.

\vspace{-0.07em}
\subsection{Main results}
\tabref{tab:experiments:plug_and_play} shows our study on the effect of illumination change to existing depth completion methods. As expected, models pretrained on KITTI are unable to generalize to nuScenes because of the photometric domain gap; errors in night time are generally higher than daytime. On its own, SpaDe pretrained on VKITTI (row 1) improves over \textbf{all KITTI and Waymo pretrained} models, despite not using the RGB image. This shows the robustness of using sparse range as an additional modality to enable transfer of model across various lighting conditions. Augmenting pretrained models with SpaDe (rows with ``+SpaDe'') significantly improves results across all metrics in both day and nighttime splits. We additionally trained baselines with Waymo and show that the results hold. This promising result illustrates the plug-and-play capability of SpaDe, which can improve domain generalization while being agnostic to model architecture. 

Moreover, all models can easily be improved with future iterations of SpaDe. To demonstrate this, we train SpaDe-V2 on VKITTI and Synthia. We observed consistent improvement when augmented both KITTI and Waymo pretrained models with SpaDe. With a better SpaDe model, downstream models also improves to similar degree, implying seamless integration with updates to SpaDe, i.e., forward-compatibility.

In \tabref{tab:experiments:outdoor}, we compare with the current trend of using image-to-image translation to re-balance day and nighttime distributions (and to increase the data volume). 
We observe a positive influence from training the baselines on translated images as compared to those on the original nuScenes with consistent improvement in both day and night time results. When compared to our method, the results were surprising. 
Even without URL, if one were to train the downstream model with SpaDe frozen and augmented to process the sparse inputs, there are immediate benefits. This is thanks to the inductive bias coming from the dense depth produced by SpaDe.
Additionally, as typical convolutions are not suited for processing sparse inputs \cite{wong2020unsupervised}, providing the depth completion model with dense, albeit an approximation, depth allows the network to properly make use of convolution operations.
This is highlighted in MSG-CHN, where we improve the nuScenes model from an MAE of 1713.83 to 1387.85. Moreover, when evaluated on nighttime scenes, all networks augmented with SpaDe received improvements. In fact, without even training on the additional translated images, models augmented with SpaDe are competitive and sometimes even improve over those that were. 
We also note that image translation requires a large computation overhead during training.

Finally, we compare URL against those trained with image-to-image translation (expanding dataset) and nighttime enhancement. URL improves pretrained baselines by 11.58\% in all-day scenarios, (11.4\% day, 12.61\% night) and 11.65\% on across \textit{all models} on all-day (11.23\% day, 13.12\% night). We improve over enhancement, by 10.2\% (9\% day, 15.4\% night) and over translation by 13.1\% (13.3\% day, 9.9\% night). 

We attribute such performance to the proposed uncertainty-driven residual learning (URL) scheme. While enhancement module attempts to close the domain gap, the possibility of introducing artifacts arises under dynamic illumination conditions. On the contrary, SpaDe operates on the illumination-robust modality, with potential erroneous regions in prediction quantified by predictive uncertainty. URL, in turn, utilizes adaptive weighting to preserve high confidence regions, re-purposing model capacity to learn the residual. Leveraging SpaDe as a strong depth prior or regularizer, URL acts as a validator given the image, correcting more uncertain regions, which improves over the use of translation or enhancement.

Our method improves over low-illumination region in \figref{fig:method_overview}(a), depth discontinuity in \figref{fig:method_overview}(c), and missing sparse depth regions in \figref{fig:method_overview}(b).
This demonstrates the inductive bias from SpaDE enables a robust estimation under low-illuminated conditions and homogeneous regions. 

\vspace{-0mm}
\section{Discussion}
We have proposed a multimodal fusion method for all-day depth completion, which leverages the strength of complementary sensor configuration under diverse illumination conditions. SpaDe, trained on readily available synthetic data, utilizes sparse range to approximate dense depth and their predictive uncertainty.  SpaDe can be used plug-and-play without training, and forward compatibility allows seamless integration of improved SpaDe models to boost performance. Given the target dataset, URL offers improved performance compared to existing methods that rely on image translation techniques, which are prone to introducing artifacts in the training images; training on them may backfire. 

Nonetheless, our method does have several limitations. Given the case of high uncertainty in approximated depth and low-illumination in nighttime images, model estimates from URL are not informative. This can be partially mitigated by including predictive uncertainty in downstream models. Our work focuses on ease of use for SpaDe and its applicability in plug-and-play; we leave design of downstream models to future works.
Another avenue is to enhance image capture, where computational imaging is relevant. This may bridge the gap between daytime and nighttime images, lending to a more sensor driven paradigm. As our approach is the first all-day depth completion method, we will release code including data setup and processing pipelines and models; we hope our promising results will motivate innovations in enabling robust estimation under diverse illumination settings.

{\small
\small
\bibliographystyle{IEEEtran}
\bibliography{bib}
}

\end{document}